\documentclass[twoside,11pt]{article}
\usepackage{jair, theapa, rawfonts}
\usepackage{graphicx}
\usepackage{amssymb}
\usepackage{subfigure}

\begin{document}

\title{Improving Skip-Gram based Graph Embeddings via Centrality-Weighted Sampling}

\author{\name Pedro Almagro Blanco \email pedro.almagro@ub.edu \\
\addr 
	 Universitat de Barcelona Institute of Complex Systems (UBICS)\\ 
	 Departament de F\'isica de la Mat\`eria Condensada\\
	 Universitat de Barcelona, Martı\'i Franqu\`es 1, E-08028 Barcelona, Spain.\\
		Complex Systems Modelling Group\\
       Universidad Central de Ecuador\\
       Quito, 170129, Ecuador.
       \AND
       \name Fernando Sancho Caparrini \email fsancho@us.es \\
       \addr Dpt. Computer Science and Artificial Intelligence\\
       University of Sevilla\\
       Sevilla, 41012, Spain.}

\maketitle

\begin{abstract}
Network embedding techniques inspired by \textit{word2vec} represent an effective unsupervised relational learning model. Commonly, by means of a \textit{Skip-Gram} procedure, these techniques learn low dimensional vector representations of the nodes in a graph by sampling \textit{node-context} examples. Although many ways of sampling the \textit{context} of a node have been proposed, the effects of the way a \textit{node} is chosen have not been analyzed in depth. To fill this gap, we have re-implemented the main four \textit{word2vec} inspired graph embedding techniques under the same framework and analyzed how different sampling distributions affects embeddings performance when tested in node classification problems. We present a set of experiments on different well known real data sets that show how the use of popular centrality distributions in sampling leads to improvements, obtaining speeds of up to 2 times in learning times and increasing accuracy in all cases. 

\end{abstract}

\section{Introduction}

The application of neural encoders to texts has provided very interesting results. In 2013, T. Mikolov et al. \cite{eff} presented two architectures, under the generic name of \emph{word2vec}, minimizing computational complexity of word representation while maintaining grammatical properties present in the texts from which the words are extracted: Continuous bag-of-words (CBOW), and Skip-gram. In them a \emph{context of a word} in a text is defined as the set of words that appear in its adjacent positions, and both architectures consist of feed-fordward artificial neural networks with three layers: an input layer, a hidden (encoding) layer  and an output layer, but they differ in the objective function they try to approximate. On the one hand, CBOW architecture receives the context of a given word as input and tries to predict that word as output. On the other hand, Skip-gram architecture receives the word as input and tries to predict the context associated with it. The main objective of the work of Mikolov et al. is to reduce the complexity in the neural model allowing the system to learn from a large volume of textual data. Through the relationship established between vocabulary words and their contexts, the model captures different types of similarity, both functional and structural, and provides an embedding of words in vector space that reflects these similarities \cite{lreg}. In the case of Skip-Gram architecture, two main optimizations have been presented: Negative Sampling, that modifies only some of the weights related to negative examples (words not in \textit{context}), and Hierarchical Softmax, where the output vector is determined by a tree-like traversal of the network layers \cite{w2v}.

Graph data appears in a number of application domains such as chemistry, social sciences, and physics, where logical problems commonly addressed are node classification \cite{neville2000iterative}, link prediction \cite{Liben-Nowell:2007:LPS:1241540.1241551}, and network representation learning \cite{zhang2018network}, to name but a few. Many successful methods inspired by language modeling have been developed recently for graph embedding. Those methods allow to obtain vector representations of nodes in a low dimensional space through sampling the relations between them in a graph.  

In this work, we evaluate the use of centrality measures to improve efficiency of four of the most popular \textit{word2vec} inspired graph embedding techniques: DeepWalk \cite{deepwalk}, LINE \cite{Tang:2015:LLI:2736277.2741093}, node2vec \cite{grover2016node2vec} and Neighborhood Based Node Embeddings (NBNE) \cite{pimentel2018fast}. We analyze the previous four models using five different centrality measures, and we obtain some important conclusions: (1) in all cases, centrality-weighted sampling speeds up convergence (x2 in some cases) of node classification tasks and, (2) there is a fixed ranking in the goodness of centralities when used in this context. This conclusions have been obtained from comprehensive experiments over two well-known datasets. Obtained results present a new application for node centrality measures to improve the efficiency of language modeling based graph embedding techniques.

The rest of the paper is arranged as follows. In Section \ref{LM} we summarize the fundamentals of language modeling through Skip-Gram model. The four graph embedding techniques under evaluation will be presented in Section \ref{GET}. In Section \ref{M} we present our approach to improve efficiency of Skip-Gram based graph embedding techniques. Section \ref{ED} is devoted to outline our experiments, and their results are presented in Section \ref{E}. In Section \ref{RW} we present works related to this one presented here. We close with our conclusions in Section \ref{C}.

\section{Language Modeling with Skip-Gram}
\label{LM}

In the following, we will present some basics related to language modeling necessary to describe the implementation that we have carried out of the four methods under study. In general terms, the goal of language modeling is to estimate the likelihood of a specific sequence of words appearing in a corpus. More formally, given a sequence of words $w_1 , w_2 , ... , w_n$, where $w_i \in \mathcal{W}$ ($\mathcal{W}$ is the vocabulary), we would like to maximize: 
$$\Pr(w_n |w_1 , w_2 , ... , w_{n - 1} )$$
over all the training corpus. 

As mentioned above, recent works have focused on using probabilistic neural networks to build general representations of words. The goal is to learn a latent representation, $\phi : \mathcal{W} \rightarrow \mathbb{R}^{|\mathcal{W}|\times d}$. This mapping $\phi$ represents the $d$-dimensional latent representation associated with each word $w$ in the vocabulary.

Skip-Gram is a language model that, from a corpus of texts, maximizes the co-occurrence probability among the words that appear within a context (of prefixed size, $c$) in a sentence of the corpus. First, instead of using the context to predict a missing word, it uses one word to predict the context. Secondly, the context is composed of the words appearing to both the right and left of the given word in the sentence. Finally, it removes the ordering constraint on the problem, requiring the model to maximize the probability of any word appearing in the context regardless of its offset from the given word. More formally, given a sequence of training words $w_1 , w_2 , ... , w_n$, the objective of the Skip-Gram model is to maximize the average log probability:

$$\frac{1}{n} \sum_{i=1}^n \sum_{-c \leq j \leq c, j \neq 0} \log \Pr(w_{i+j} \mid w_i) $$

As we will see later, these relaxations are particularly desirable for graph representation learning: the order independence assumption better captures the sense of \textit{neighbourhood} as it is provided in graphs; moreover, this fact will be quite useful for speeding up the training time by building small models giving one vertex at a time. 

The basic Skip-Gram formulation defines $\Pr(w_{i+j} \mid w_i)$ using the soft-max function as:

$$ \Pr(w_{i+j} \mid w_i ) = \frac{\exp(\phi(w_{i+j})'^T \phi(w_i))}{\sum_{w=1}^W \exp(\phi(w)'^T \phi(w_i))}$$

\noindent where $\phi(w)$ and $\phi(w)'$ are the \textit{input} and \textit{output} vector representations of $w$, respectively. Optimizing this model by gradient descent means taking a training example and adjusting all the parameters of the model. In other words, each training example will tweak all of the parameters in the model. Usually, the size of the vocabulary means that Skip-Gram model has a tremendous number of parameters, all of which would be updated by every one of the many training examples.  To speed up the training time, Skip-Gram authors presented two approximations: Negative Sampling and Hierarchical Soft-max \cite{mikolov2013distributed}. 

\subsection{Negative Sampling}
\label{ns}
Negative Sampling faces the cost of calculating $ \Pr(w_{i+j} \mid w_i)$ by allowing each training example to only modify a small percentage of the parameters, rather than all of them. For each observed pair $(w_{i+j},w_i)$ we sample $k$ \textit{negative} context words $w \in \mathcal{W}$ from a noise distribution $P_n(w)$.

$$\log \sigma (\phi(w_{i+j})'^T \phi(w_i)) - \sum_{p=1}^k \mathit{E}_{w_p} \sim  P_n ( w ) [\log \sigma (\phi(w_p)'^T \phi(w_i))] $$

\noindent which is used to replace every $\log \Pr(w_{i+j} \mid w_i)$ term in the Skip-Gram objective. Thus the task is to distinguish the correct samples from negative ones obtained from the noise distribution $P_n(w)$, where there are $k$ negative samples for each positive data sample. 

\subsection{Hierarchical Soft-max}


Other way to deal with the high costs of calculating $ \Pr(w_{i+j} \mid w_i)$ is Hierarchical Softmax. This model factorizes the conditional probability assigning the words to the leaves of a binary tree, turning the prediction problem into maximizing the probability of a specific path in the hierarchy. If the path to word $w_{i+j}$ is identified by a sequence of tree nodes, ($b_0 = \textrm{root},\ b_1,\ b_2,\ ... ,\ b_{log|\mathcal{W}|= w_ {i+j} })$, then:

$$ \Pr(w_{i+j} \mid w_i) = \prod_{l=1}^{log|\mathcal{W}|} \Pr(b_l | w_i )$$

Now, $\Pr(b_l | w_i)$ could be modeled by a binary classifier that is assigned to the parent of the node $b_l$ as next equation shows:

$$\Pr(b_l \mid w_i ) = \sigma(\phi(w_i )^T \alpha(b_l))$$ 

where $\sigma(x) = 1/(1 + \exp(-x))$ and $\alpha(b_l ) \in \mathbb{R}^d$ is the representation assigned to tree node $b_l$ 's parent. This reduces the computational complexity of calculating $\Pr(w_{i+j} | w_i )$ from $O(|\mathcal{W} |)$ to $O(log |\mathcal{W}|)$. Also, unlike the standard soft-max formulation of the Skip-Gram which assigns two representations $\phi(w)$ and $\phi'(w)$ to each word $w$, the hierarchical soft-max formulation has one representation $\phi(w)$ for each word $w$ and one representation $\alpha(b_l)$ for every inner parent node of $b_l$ in the binary tree. 

\section{Graph Embedding Techniques Inspired by Language Modeling}
\label{GET}

Recent work in graph embedding uses Skip-Gram probabilistic neural networks to build general representations of the nodes in a graph. In this work we analyze how centrality measures can be used to improve efficiency of the four graph embedding techniques presented below. As we don't work with weights nor attributes, we will not describe details of the techniques related with that aspects.

\subsection{DeepWalk}

\textit{DeepWalk} is a generalization of language modeling that explores a graph $G=(V,E)$ through a stream of short random walks and learns representation of nodes using a Skip-Gram architecture \cite{deepwalk}. These walks can be thought of as short sentences and phrases in a special language; the direct analog is to estimate, given a random walk $v_1 , v_2 , v_3 , ... , v_n$, the likelihood of observing vertex $v_{i+j}$ given a vertex $v_i$ in the random walk, i.e.

$$\Pr(v_{i+j} | v_i ) = \frac{\exp(\psi(v_{i+j})'^T \psi(v_i))}{\sum_{v=1}^V \exp(\phi(v)'^T \phi(v_i))}$$

The goal is to learn a latent representation, $\psi : V \rightarrow \mathbb{R}^{|V|\times d}$. This mapping $\psi$ represents the $d$-dimensional latent representation associated with each vertex $v$ in the graph. The objective of the Skip-gram model is to maximize the average log probability:

$$\frac{1}{n} \sum_{i=1}^n \sum_{-c \leq j \leq c, j \neq 0} \log \Pr(v_{i+j} \mid v_i) $$

where $c$ represents context size. DeepWalk applies a Skip-Gram model with Hierarchical Soft-max to random walks formulating a method which generates low-dimensional representations of networks in a continuous vector space. 

\subsection{LINE}

LINE (Large-scale Information Network Embedding) also uses a similar Skip-Gram architecture to embed the networks but it differs in some aspects from DeepWalk. First, LINE don't use random walks to generate examples, however it uses solely the neighbor information of every node to preserving both the first-order and second-order proximity. LINE model trains the first-order proximity and second-order proximity separately and then concatenate the obtained embeddings for each vertex \cite{Tang:2015:LLI:2736277.2741093}. The first-order proximity tries to maximize for each edge $(v_i, v_j) \in E$, the joint probability:

$$ \Pr(v_i \mid v_j ) = \sigma(\psi(v_i)^T \psi(v_j))$$

As it uses the same \textit{input} and \textit{output} vector representations, first-order proximity can deal only with undirected graphs. To avoid the trivial solution $\psi(v_{ik}) = \infty$, for $i=1, . . . , |V|$ and $k = 1, . . . , d$, the first-order proximity is approximated using a Negative Sampling architecture. The second-order proximity is also learned using Negative Sampling but using different \textit{input} and \textit{output} vector representations (as presented in Section \ref{ns}). In both cases positive examples are formed by a node and one of his neighbors and negative examples are extracted from a modified unigram distribution $P_n(v) \propto d_v^{3/4}$ as proposed in \cite{mikolov2013distributed}, where $d_v$ is the out-degree of vertex $v$.

\subsection{node2vec}
\label{n2v}
node2vec is a generalization of DeepWalk in which random walks are guided by two parameters $p$ and $q$. Given a random walk that has just crossed the edge $(t,v)\in E$, the probability of taking the edge $(v,x)\in E$ corresponds to:
\newcommand{\threepartdef}[6]
{
	\left\{
		\begin{array}{lll}
			#1 & \mbox{if } #2 \\
			#3 & \mbox{if } #4 \\
			#5 & \mbox{if } #6
		\end{array}
	\right.
}

$$\gamma_{pq}(v,x) =  \threepartdef
{\frac{1}{p}}      {dis(t,x)=0}
{1}      {dis(t,x)=1}
{\frac{1}{q}} {dis(t,x)=2} $$

where $dis(t,x)$ represents the length of the minimum path between node $t$ and node $x$. node2vec is identical to DeepWalk with the exception that it explores new methods to generate random walks, at the cost of introducing more hyperparamenters \cite{grover2016node2vec}.

\subsection{Neighborhood Based Node Embedding}

Neighborhood Based Node Embedding (NBNE) uses a Skip-Gram model with Negative Sampling to learn representations of nodes in a graph from positive examples formed by a node and one of his neighbors and negative examples extracted from a modified unigram distribution $P_n(v) \propto d_v^{3/4}$. NBNE is equivalent to second-order proximity in LINE, its performance is lower than previous models but its training is faster due to the simplicity of its algorithm \cite{pimentel2018fast}.

\section{Method}
\label{M}

Under the premise that more central nodes are more informative when learning representations for later classification, we have implemented the four methods under the same Negative Sampling architecture and we have introduced $\lambda_i$ (the prestige of vertex $v_i$) in the conditional probability.

\subsection{Overview}

All graph embedding techniques under analysis requires a set of positive examples and a set of negative examples. Negative examples will be drawn from the uniform distribution. Positive examples will be generated in a different manner for each case: DeepWalk and node2vec consider a set of short truncated random walks as corpus from which to extract positive examples while LINE and NBNE generate positive examples using the neighbors of a node. 

\subsection{Algorithm}

To increase efficiency, we decide to use one negative example for each positive example ($k=1$). Then, in our framework, the term $\log\Pr(v_{i+j} \mid v_i)$ for DeepWalk and node2vec objective function becomes:

$$\log \lambda_i \sigma (\psi(v_{i+j})'^T \psi(v_i)) - \mathit{E}_{v_p} \sim  P_n ( v ) [\log \sigma (\psi(v_p)'^T \psi(v_i))] $$

In the case of DeepWalk, $v_{i+j}$ is obtained from random walks with a context of size $c=30$. In the case of node2vec, $v_{i+j}$ is obtained from modified random walks as described in Section \ref{n2v} and also with a context of size $c=30$. The term $\log\Pr(v_j,v_i)$ to be maximized in  NBNE becomes

$$\log \lambda_i \sigma (\psi(v_{j})'^T \psi(v_i)) - \mathit{E}_{v_p} \sim  P_n ( v ) [\log \sigma (\psi(v_p)'^T \psi(v_i))] $$

for each edge $(v_i, v_j) \in E$. In the case of LINE, we have approximated the \textit{first-order proximity} by using NBNE and \textit{second-order proximity} by selecting nodes at distance 2 as positive examples. The presented framework allows to compare different embedding approaches under a common framework taking advantage of high parallelism. 

\subsection{Centrality-weighted Sampling}
\label{CWS}

Next we describe the centrality-weighted sampling treatment which improves the effectiveness of the embeddings when facing node classification tasks. We will use centrality information of node $v_i$ to determine parameter $\lambda_i$. Centrality measures under consideration are Degree, Betweenness Centrality (BC) \cite{freeman}, Closeness Centrality (Clos) \cite{bavelas1950communication}, PageRank (PR) \cite{pr} and Load Centrality  \cite{brandes2008variants}. As we will show in next sections, efficiency of all techniques under analysis is improved using proposed sampling techniques. 

\section{Experimental Design}
\label{ED}

To train the models, we used 1M positive examples (and 1M negative examples) and 200 dimensions (as in original methods). A single learning process iterates through all positive and negative examples with a batch size of 100k. We used the same random initial weights for every centrality measure following \cite{grover2016node2vec}. Stochastic gradient descent (SGD) \cite{NIPS2011_4390} is used in our experiments to optimize free parameters. The derivatives are estimated using the back-propagation algorithm and the learning rate for SGD is 0.1\%. Our framework was implemented with the \textit{TensorFlow} \cite{45381} wrapper \textit{Keras} \cite{chollet2015keras}. We used the logistic regression classifier from \textit{LibLinear} \cite{REF08a}. All experiments were run on hardware with 32GB RAM, a single 3.4 GHz CPU, and two GeForce GTX 1080 GPUs. 

\subsection{Datasets}

We evaluate the influence of centrality weighted sampling with the following two commonly used benchmark data sets: Cora \cite{T8/HUIG48_2017} and Citeseer \cite{Giles98citeseer:an}. Cora and Citeseer are citation networks where nodes represent documents and links represent citations. In both cases, class labels represent the main topic of the document each node have exactly one class label. Statistics about data sets are summarized in Table 1.

\begin{table}[]
\label{ds}
\center
\begin{tabular}{lrrr}
\hline
\multicolumn{1}{c}{Dataset} & \multicolumn{1}{c}{$|V|$} & \multicolumn{1}{c}{$|E|$} & \multicolumn{1}{c}{\#classes} \\ \hline
Cora                        & 2,708                    & 5,429                    & 7                             \\
Citeseer                    & 3,327                       &  4,732                       & 6                             \\ \hline
\end{tabular}
\caption{Dataset statistics.}
\end{table}

\section{Experiments}
\label{E}

Let us perform empirical evaluations with the objective of analyze how centrality measures can help to improve the efficiency of presented graph embedding methods when facing node classification. 

\begin{figure}
\label{repr}
\centering
	\fbox{\subfigure{\includegraphics[scale=0.5]{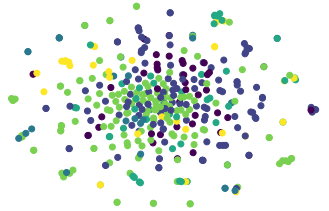}}}
	\fbox{\subfigure{\includegraphics[scale=0.493]{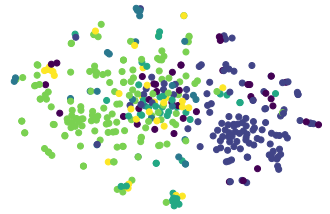}}}
	\caption{Visualization of a fraction of \textit{CiteSeer} network. Documents are mapped to the 2-D space using the t-SNE package (only 400k samples). (right) Embedding achieved with node2vec+BC, embedding achieved with node2vec baseline.}
\end{figure}

In the label classification setting, every node is assigned one label from a finite set. During the training phase, we observe the representation of a certain fraction of nodes and their labels. The task is to predict the labels for the remaining nodes. Each experiment has been repeated 4 times, obtaining a standard deviation smaller than 0,006\% in all cases. In Figure \ref{repr} we show two 2-D representations of node representations achieved with node2vec over CiteSeer data (color of a node indicates the topic of the document). Embedding at the right side have been achieved with Betweenness Centrality weighted sampling and embedding at the left side without weighted sampling. Both embeddings have been achieved training only with 400k samples.

\subsection{Cora} 

Experiments with Cora data set reveal that the use of centrality measures when sampling nodes to participates as positive examples in the different graph embedding methods presented in Section \ref{GET} leads to a representative speedup in the convergence to an optimal embedding for node classification.

\begin{table}[]
\label{mf1cora}
\center
\begin{tabular}{lcccc}
\hline
Metric & DeepWalk        & LINE            & node2vec        & NBNE            \\ \hline
BC     & \textbf{0.6338} & \textbf{0.5386} & \textbf{0.6337} & \textbf{0.5354}          \\
Load   & 0.6329          & 0.5375          & 0.6330          & 0.5354 \\
Deg    & 0.6380          & 0.5206          & 0.6375          & 0.5195          \\
PR     & 0.6330          & 0.5162          & 0.6373          & 0.5116          \\
Clos   & 0.5932          & 0.4912          & 0.5858          & 0.4876          \\
Base   & 0.5805          & 0.4848          & 0.5952          & 0.4667          \\ \hline
\end{tabular}
\caption{\textit{Micro-F1} for node classification in \textit{Cora} data set.}
\end{table}

The results for this data set are listed in Table 2. We include the baseline and the centrality-weighted results. node2vec and DeepWalk significantly outperforms all existing approaches on Cora data set. Both obtain the highest improvement with BC but their results are really similar to those obtained using Load. Also, we can mention that node2vec and DeepWalk do not take full advantage of PageRank centrality. Another fact that we can observe in this table is the clear ranking between different centrality measures. Betweenness Centrality ranks first, next Load, next Degree, next PageRank and at the last place we find Closeness Centrality. Betweenness and Load have similar behaviors. It can be due to the similarity between those two metrics. 

\begin{figure}
\label{mf1coraf}
	\begin{center}
		\includegraphics[scale=0.41]{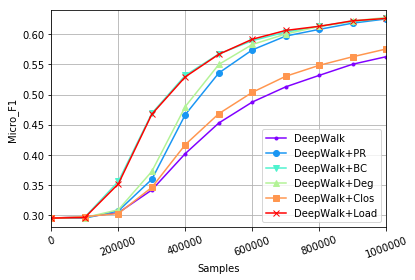}
		\includegraphics[scale=0.41]{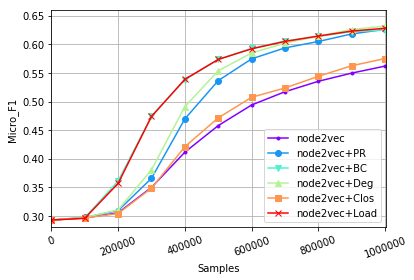}
		\includegraphics[scale=0.41]{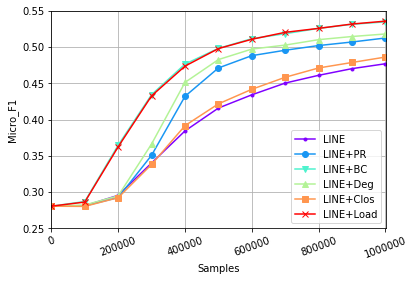}
		\includegraphics[scale=0.41]{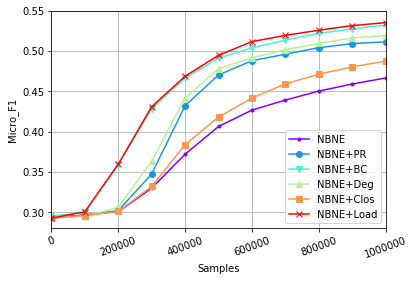}
	\end{center}
	\caption{\textit{Micro-F1} for Cora label classification problem using baselines and centrality weighted sampling models.}
\end{figure}

Figure 2 presents the relation bwetween Micro-F1 metric and the number of examples used to obtain the embedding in Cora data set for baseline methods and using centrality-weighted samplings. As we can observe, the highest improvement is produced between 200k and 600k samples for the Cora data set. In addition, we confirm really similar behavior for Betweenness Centrality and Load Centrality, for Degree and PageRank and for baseline and Closeness Centrality.

\subsection{Citeseer} 

Experiments with Citeseer data set reveal similar insights. Again, the use of centrality measures when sampling nodes to participate as positive examples in the different graph embedding methods presented in Section \ref{GET} leads to a representative speedup in the convergence to an optimal embedding for node classification.

\begin{table}[]
\label{mf1citeseer}
\center
\begin{tabular}{lcccc}
\hline
Metric & DeepWalk        & LINE            & node2vec        & NBNE            \\ \hline
BC     & 0.4525 & \textbf{0.4222} & \textbf{0.4556} & 0.4240          \\
Load   & \textbf{0.4536}          & 0.4215          & 0.4542          & \textbf{0.4249} \\
Deg    & 0.4468          & 0.3977          & 0.4483          & 0.4001          \\
PR     & 0.4448          & 0.3911          & 0.4398          & 0.3929          \\
Clos   & 0.4234          & 0.3815          & 0.4280          & 0.3863          \\
Base   & 0.4019          & 0.3653          & 0.4061          & 0.3616          \\ \hline
\end{tabular}
\caption{\textit{Micro-F1} for node classification in \textit{Citeseer} data set.}
\end{table}

The results for Citeseer are listed in Table 3. node2vec and DeepWalk significantly outperforms all presented models on Citeseer classification problem. In this case, node2vec has highest improvement with BC but DeepWalk with Load. node2vec and DeepWalk take advantage of PageRank on CiteSeer data set. The centrality ranking remains so similar than in the Cora case.

\begin{figure}
\label{mf1citeseerf}
	\begin{center}
		\includegraphics[scale=0.41]{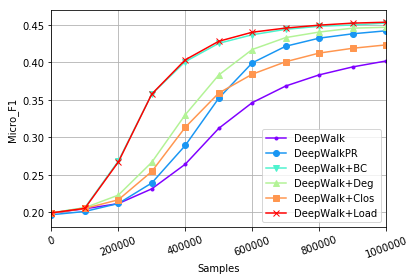}
		\includegraphics[scale=0.41]{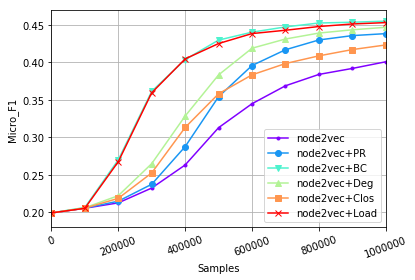}
		\includegraphics[scale=0.41]{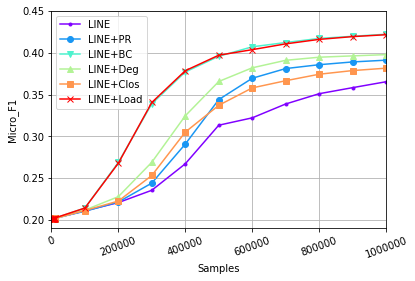}
		\includegraphics[scale=0.41]{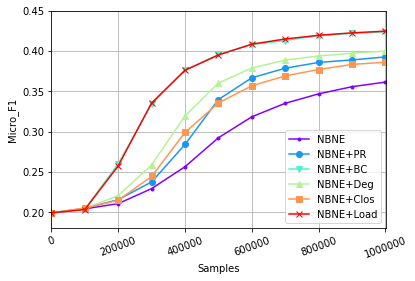}
	\end{center}
	\caption{Micro-F1 for Citeseer label classification problem using different centrality measures when sampling and baseline models.}
\end{figure}

Figure 3 present the relation bwetween Micro-F1 metric and the number of examples used to obtain the embedding of the different base models and their centrality weighted versions for Citeseer data set. As we can observe, the highest improvement is produced again between 200k and 600k samples. We confirm almost the same behavior for Betweenness and Load Centrality in all methods. For node2vec model, Load Centrality performs a little bit better for a big number of examples (close to one million). In this case, Closeness Centrality performs better than on Cora, performing better than PageRank version at the very initial phase (under 500k examples).

\section{Related Works}
\label{RW}

Usually, formalizations of graphs include only positive relation instances, leaving the door open for a variety of methods for selecting negative examples. In \cite{DBLP:journals/corr/abs-1708-06816} the authors present an empirical study on the impact of Negative Sampling on the learned embeddings, assessed through the task of link prediction. They focus in compare well known methods for Negative Sampling (random generation and corrupting positive examples) and propose two new embedding based sampling methods.

In the work titled \textit{Robust negative sampling for network embedding} the authors provide theoretical arguments that reveal how Negative Sampling can fail to properly estimate the Skip-Gram objective, and why it is not a suitable candidate for the network embedding problem \cite{armandpour2019robust}. They show that Negative Sampling can learn undesirable embeddings, as the result of the \textit{Popular Neighbor Problem}. This deviation of Negative Sampling from that ideal behavior is mainly caused by allowing a node to choose its neighbor as a negative sample. This problem is more severe when high-degree nodes are present. They present a new method that alleviates this problem by using a new negative sampling scheme and penalization of the embeddings.

In \cite{Wang2014KnowledgeGE} the authors present an optimization of Negative Sampling in the case of embeddings working with relational databases \cite{transe,Wang2014KnowledgeGE}. They tend to give more chance to replacing the head entity of a relation if it is one-to-many and more chance to replacing the tail entity if the relation is many-to-one. In this way, the chance of generating false negative labels is reduced. 

As we shown, some work regarding to the methods selecting negative samples in Skip-Gram graph embedding techniques have been presented. Only LINE model introduces a parameter in the objective function to guide positive examples selection. But LINE authors did not make any analysis of the influence of this parameter in the efficiency of the method and did not present any advantage of this approach. Our work fills this gap presenting a detailed study of such influence not only over LINE but over other similar graph embedding techniques.

\section{Conclusions} 
\label{C}

The use of probability distributions in the selection of positive examples when using graph embedding methods based on Skip-Gram allows to obtain a higher performance in node classification tasks. Both the methods based on random paths and those that construct the positive examples only from the vicinity of the nodes have demonstrated a significant improvement in efficiency by using distributions related to centrality measures in the nodes of the graph. From our knowledge, this work represents the first analysis of this Skip-Gram modification on graphs. Experiments on real data illustrate the effectiveness of our approach on challenging label classification tasks. Our results show that we can create fast and scalable meaningful representations for large graphs making use of centrality measures when selecting positive examples. Our method significantly outperforms other methods designed for the same purpose. 

Starting from the premise that the centrality of a node in a network is a sign that this node is more informative when performing an embedding, in this work we have presented an analysis concerning some of the most popular measures of centrality. The results show a significant improvement in all methods under study. Specifically, centralities not based on random paths such as Betweenness Centrality have demonstrated their usefulness to accelerate embedding proceses. In addition, the results show a clear ranking in the centralities according to their goodness. Experiments with a bigger number of examples and with other centrality measures, datasets and methods are still pending. Experiments regarding other rekational learning tasks as link prediction should be also considered. Another worth to mention future line of research consists on applying centrality-based distributions to negative examples.

One of the most interesting conclusions that can be drawn from this study is derived from the fact that Betweenness and Load centralities are the most successful measures in the presented context. From our point of view, this is because the information that these two measures provide is different from the information contained in the random paths and in the first and second order proximities in which models under study are based. Both centralities contain information on minimum paths not explicitly explored by methods under analysis and we believe that this fact is key to understand the improvement presented.

\acks{The work was supported in part by the GLVEZUS Project of Universidad Central de Ecuador, by the project ``Metodolog\'{\i}as de datos aplicadas al an\'alisis de las exposiciones art\'{\i}sticas en Andaluc\'{\i}a para el desarrollo de la econom\'{\i}a creativa", from Fundaci\'on P\'ublica Centro de Estudios Andaluces (2017-2019) and by the projects TIN2013-41086-P (LOCOCIDA Project) and FIS2016-76830-C2-2-P (Adaptabilidad y Cooperación en Sistemas Biosociales en la Multiescala II) from Ministerio de Econom\'{\i}a, Industria y Competitivad of Spain (FEDER funds from EU).
}

\vskip 0.2in
\bibliography{biblio}
\bibliographystyle{theapa}

\end{document}